\documentclass[conference,compsoc]{IEEEtran}

\IEEEoverridecommandlockouts                              

\usepackage[utf8]{inputenc}
\usepackage[T1]{fontenc}
\usepackage{graphicx}

\usepackage{graphics} 
\usepackage{epsfig} 
\usepackage{mathptmx} 
\usepackage{mathptmx} 
\usepackage{amsmath} 
\usepackage{amssymb}  
\usepackage{float}
\usepackage{cite}
\usepackage{ragged2e} 

\usepackage{subfig}
\usepackage{chngcntr}
\usepackage{xcolor}
\counterwithin{figure}{section}
\usepackage{indentfirst}
\usepackage{siunitx}
\usepackage[margin=1.0in]{geometry}
\usepackage{algorithm,algorithmic,amsmath}
\usepackage{enumitem}
\usepackage{hyperref}
\usepackage{breakurl}

\title{\LARGE \bf
Optimization of XNOR Convolution for Binary Convolutional Neural Networks on GPU 
}

\author{
\IEEEauthorblockN{Mete C. Kaya}
\IEEEauthorblockA{Electrical and Electronics Engineering,\\Middle East Technical University\\e209405@metu.edu.tr}
\and
\IEEEauthorblockN{Alperen İnci}
\IEEEauthorblockA{Graduate School of Informatics,\\ Middle East Technical University\\alperen.inci@metu.edu.tr}

\and
\IEEEauthorblockN{Alptekin Temizel}
\thanks{Submitted for Consideration for Publication in High Performance Computing Conference 2020}\\
\IEEEauthorblockA{Graduate School of Informatics,\\Middle East Technical University\\atemizel@metu.edu.tr}
}

\begin{document}

\maketitle

\thispagestyle{empty}
\pagestyle{empty}

\begin{abstract}

Binary convolutional networks have lower computational load and lower memory foot-print compared to their full-precision counterparts. So, they are a feasible alternative for the deployment of computer vision applications on limited capacity embedded devices. Once trained on less resource-constrained computational environments, they can be deployed for real-time inference on such devices. In this study, we propose an implementation of binary convolutional network inference on GPU by focusing on optimization of XNOR convolution. Experimental results show that using GPU can provide a speed-up of up to $\mathbf{42.61\times}$ with a kernel size of $\mathbf{3\times3}$. The implementation is publicly available at \url{https://github.com/metcan/Binary-Convolutional-Neural-Network-Inference-on-GPU}.

\end{abstract}

\section{Introduction}
Deploying deeper neural networks having large number of parameters have been commonplace in the recent years. While this led to state-of-the-art performance, it also came with high computational cost and memory requirements, which has limited their deployment on lower capacity devices. For this reason, there has been an increased interest in development of efficient deep neural network models that can work effectively on devices with limited capabilities. Two fundamental approaches aiming to solve this problem are; designing of smaller Convolutional Neural Network (CNN) models, and pruning existing networks to obtain smaller networks with comparable performance such as MobileNetV2 \cite{Mobilnetv2}, EfficientNet \cite{EfficentNet} and ShuffleNet \cite{ShuffleNet}. On the other hand, Binary-Weight-Networks provides a distinct alternative approach to this problem where full-precision operations are replaced with binary-precision operations. The main benefit of these models is that both memory and computation requirements are significantly reduced without changing the parameter size. Even though this comes with a performance penalty, it allows a trade-off between network performance and computational complexity to run such networks on limited capability devices.
\\
In the recent years, an increasing number of binary network models and implementations have been proposed \cite{BNSurvey}. BitFlow \cite{Bitflow} is reported to have $1.8\times$ speedup against standard binary network implementations, while it has a $11.5\times$ speedup against  full-precision networks. XNOR-SRAM \cite{XNOR-SRAM} is a hardware solution for ternary-XNOR-and-accumulate (XAC) operations, exhibiting $33\times$ energy saving. XNOR-Net \cite{xnor}, a prominent type of binary network, has been reported to have $32\times$ memory saving and $62.7\times$ theoretical speed-up on CPU. XNOR-Net++ \cite{XNOR++} proposed an improved training algorithm for binary networks, achieving $6\%$ higher accuracy on ImageNet compared to XNOR-Net \cite{xnor}.  


In this paper, we propose an implementation of binary convolutional network on GPU and optimization of binary XNOR convolution.  While training of deep networks have high computational cost, training is generally done once before the deployment of the network. Hence training can be done on systems with higher computational and memory capacity. On the other hand, the inference path of the network is run continuously once it is deployed and the network is generally required to be deployed on cost-effective devices for real-life applications. Hence, inference is desired to have low computational complexity for cost-effective and widespread deployment. In this work, XNOR-Net binary network \cite{xnor} is taken as the reference method and the forward path of this algorithm, used for the inference, is optimized on GPU.

\section{Background}
\label{sec:background}
The main bottleneck of CNN models is the high-memory requirement, which hinders their deployment on limited capacity devices. Binary-Weight-Networks, \cite{xnor} binarizes the weight values as opposed to  using full-precision and can achieve $32\times$ memory saving and $2\times$ speed-up. By approximating both weights and input as binary values, X-NOR Net can achieve $58\times$ speed-up in implementation on CPUs. In this section we first describe the  binary networks in general and then describe the specifics of XNOR-Net. 

\subsection{Binary Weight Networks}
First, the weight values need to be approximated as binary values so convolution can be implemented with the help of efficient subtraction and addition operations. The binary weights, $\mathbf{B} \in\{+1,-1\}^{C \times W \times H}$, are represented by the triplet $c$, $w$, $h$, where
$w \in [0, W)$ indicates the row, $h \in [0, H)$ indicates the column, and $c \in [0, C)$ indicates the channel. The weights, $\mathbf{W} \in \mathcal{W}$, are represented as binary $\mathbf{B} \in\{+1,-1\}^{C \times W \times H}$ by the help of a scaling factor $\alpha \in \mathbb{R}^{+}$ $\mathbf{W} \approx \alpha \mathbf{B}$. Then the convolution can be approximated as in Eq. \ref{conveq} where $\oplus$ indicates a convolution without any multiplication. 

\begin{equation}
    \label{conveq}
    \mathbf{I} * \mathbf{W} \approx(\mathbf{I} \oplus \mathbf{B}) \alpha
\end{equation}

$\mathbf{B}=\mathcal{B}_{l k}$ is a binary filter and $\alpha=\mathcal{A}_{l k}$ is a scaling factor and $\mathcal{W}_{l k} \approx \mathcal{A}_{l k} \mathcal{B}_{l k}$
To find optimal solution, the optimization in Eq. \ref{OptEq} is solved.

\begin{equation}
\label{OptEq}
    \begin{gathered}
J(\mathbf{B}, \alpha)=\|\mathbf{W}-\alpha \mathbf{B}\|^{2} \\
\alpha^{*}, \mathbf{B}^{*}=\underset{\alpha, \mathbf{B}}{\operatorname{argmin}} J(\mathbf{B}, \alpha) \\
J(\mathbf{B}, \alpha)=\alpha^{2} \mathbf{B}^{\top} \mathbf{B}-2 \alpha \mathbf{W}^{\top} \mathbf{B}+\mathbf{W}^{\top} \mathbf{W}
    \end{gathered}
\end{equation}

$\mathbf{B} \in\{+1,-1\}^n$, $\mathbf{B}^{\top} \mathbf{B}$ = n and  $\mathbf{W}^{\top} \mathbf{W}$ is constant. The parameter that is to be minimized is $-2 \alpha \mathbf{W}^{\top} \mathbf{B}$ which requires maximization of $\mathbf{W}^{\top} \mathbf{B}$. Since $\mathbf{B}$ is \{+1,-1\}, the maximization can be done by taking the sign of $\mathbf{W}$ and multiplying with $\mathbf{W}^{\top}.$ By taking the derivative of $J$ with respect to $\alpha$, Eq. \ref{MaxEq} and \ref{AlphaEq} are obtained.

\begin{equation}
\label{MaxEq}
    \mathbf{B}^{*}=\underset{\mathbf{B}}{\operatorname{argmax}}\left\{\mathbf{W}^{\top} \mathbf{B}\right\} \quad \text { s.t. } \mathbf{B} \in\{+1,-1\}^{n}
\end{equation}

\begin{equation}
\label{AlphaEq}
\alpha^{*}=\frac{\mathbf{W}^{\top} \mathbf{B}^{*}}{n}
\end{equation}

By replacing $\mathbf{B}^{*}$ with $\operatorname{sign}(\mathbf{W})$, this can be written as in Eq. \ref{AlphaEq2}, which implies that optimal estimation of binary weight can be computed by taking the sign of weight and scale factor is the average of absolute weight values.

\begin{equation}
\label{AlphaEq2}
\alpha^{*}=\frac{\mathbf{W}^{\top} \operatorname{sign}(\mathbf{W})}{n}=\frac{\sum\left|\mathbf{W}_{i}\right|}{n}=\frac{1}{n}\|\mathbf{W}\|_{\ell 1}
\end{equation}

\subsection{XNOR Networks}

In addition to binarization of weights, XNOR-Network also binarizes the inputs. This can be considered as binarizing the inputs of the convolutions by the help of a binary activation function. Since both the weight and input have binary values, convolution operation can then be implemented using XNOR operation. Since both are binary vectors, convolution operation is comprised of shift and dot product operations. In the Binary Weight Network, $\mathbf{W}$ is approximated as $\alpha\mathbf{B}$ and the input $\mathbf{X}$ as $\beta\mathbf{H}$. So, it can be written that $\mathbf{X}^{\top} \mathbf{W} \approx \beta \mathbf{H}^{\top} \alpha \mathbf{B}$, $\text{ where }\ \mathbf{H}, \mathbf{B} \in\{+1,-1\}^{n}$ $\text { and }\  \alpha, \beta \in \mathbb{R}^{+}.$ This time the optimization process involves two parameters $\alpha$ and $\beta$ as in Eq. \ref{AlphaBetaEq}:

\begin{equation}
\label{AlphaBetaEq}
\alpha^{*}, \mathbf{B}^{*}, \beta^{*}, \mathbf{H}^{*} = \underset{\alpha, \mathbf{B}, \beta, \mathbf{H}}{\operatorname{argmin}}\|\mathbf{X} \odot \mathbf{W}-\beta \alpha \mathbf{H} \odot \mathbf{B}\|
\end{equation}
where $\odot$ indicates element-wise product. To put the equation into a simpler form, we can define  $\mathbf{Y} \in \mathbb{R}^{n}$ as  $\mathbf{Y}_{i}=\mathbf{X}_{i} \odot \mathbf{W}_{i}$,  $\mathbf{C} \in\{+1,-1\}^{n}$ as $\mathbf{C}_{i}=\mathbf{H}_{i} \odot \mathbf{B}_{i}$, and $\gamma \in \mathbb{R}^{+}$ as $\gamma=\beta \alpha .$ This can be written using the same approach in Binary Weight Networks as in Eq. \ref{GamaEq} and \ref{CEq}.

\begin{equation}
\label{GamaEq}
\gamma^{*}, \mathbf{C}^{*}=\underset{\gamma, \mathbf{C}}{\operatorname{argmin}}\|\mathbf{Y}-\gamma \mathbf{C}\| 
\end{equation}

\begin{equation}
\label{CEq}
\mathbf{C}^{*}=\operatorname{sign}(\mathbf{Y})=\operatorname{sign}(\mathbf{X}) \odot \operatorname{sign}(\mathbf{W})=\mathbf{H}^{*} \odot \mathbf{B}^{*}
\end{equation}

Since $\left|\mathbf{X}_{i}\right|,\left|\mathbf{W}_{i}\right|$ are independent this leads to Eq. \ref{GamaEq2}.

\begin{equation}
\label{GamaEq2}
\begin{aligned} 
\gamma^{*} & = \frac{\sum\left|\mathbf{Y}_{i}\right|}{n}=\frac{\sum\left|\mathbf{X}_{i}\right|\left|\mathbf{W}_{i}\right|}{n} \\
& \approx\left(\frac{1}{n}\|\mathbf{X}\|_{\ell 1}\right)\left(\frac{1}{n}\|\mathbf{W}\|_{\ell 1}\right) \\
& =\beta^{*} \alpha^{*}
\end{aligned}
\end{equation}
For calculating scale factors, the average of each channel is taken and convolved with 2D filter $\mathbf{k} \in \mathbb{R}^{w \times h}$. Expression and final approximation can be defined as in Eq. \ref{compK} and \ref{IWEq}.

\begin{equation}
\label{compK}
    \begin{aligned}
    \mathbf{K} &= \mathbf{A} * \mathbf{k},\quad \mathbf{A}=\frac{\sum\left|\mathbf{I}_{:, :, i}\right|}{c}, \quad \mathbf{k}_{i j}=\frac{1}{w \times h}
    \end{aligned}
\end{equation}

\begin{equation}
\label{IWEq}
\mathbf{I} * \mathbf{W} \approx(\operatorname{sign}(\mathbf{I}) \otimes \operatorname{sign}(\mathbf{W})) \odot \mathbf{K} \alpha
\end{equation}

\section{Algorithm Implementation}

\subsection{Binary Convolution}
In this section, we first describe the generic implementation of the XNOR convolution. Then, the CPU and GPU implementations and their differences over the same pipeline are described.
Binary convolution has the following steps:
\begin{enumerate}[wide, labelwidth=!, labelindent=0pt]
\item XNOR Convolution Bit operations
    \begin{enumerate}[wide, labelwidth=!, labelindent=10pt]
    \item Conversion of input data type to binary type
    \item XNOR bitwise logical operation on binary data with binary weights 
    \item Summation of output binary bits where 0 values are considered as -1. 
    \item Converting Binary to float data type.
\end{enumerate}

\item XNOR Convolution Scaling Factor Computation
    \begin{enumerate}[wide, labelwidth=!, labelindent=10pt]
    \item Channel-wise summation of input data.
    \item Multiplication of matrix $\mathbf{K}$ with the scalar $\alpha$ value.
    \end{enumerate}

\item Multiplication of float output of XNOR convolution with $\mathbf{K}$ and  $\alpha$ values.
\end{enumerate}

\subsubsection{Converting Integer to Binary}
The XNOR convolution operation is a bit-wise logical operation. The input and image tensors are stored inside registers to fully utilize the given processor. The pseudo-code is provided in Algorithm \ref{alg:inputtobinary}. 

\begin{algorithm}
  \begin{algorithmic}[1]
    \FOR{$j < IMAGE\_HEIGHT$}
      \FOR{$i < IMAGE\_WIDTH$}
        \STATE $register\_image = input\_image[j][i] >> Shift$
        \STATE $Shift += 1$\\
        \IF{ $i\bmod register_x == 7 $}
        \STATE $i = i - (kernel\_size_x - 1)/2 $ 
        \ENDIF
      \ENDFOR
        \IF{ $j\bmod register_y == 7 $}
        \STATE $j = j - (kernel\_size_y - 1)/2 $ 
        \ENDIF
    \ENDFOR
  \end{algorithmic}
   \caption{Input Image to Binary Image}
   \label{alg:inputtobinary}
\end{algorithm}

\subsubsection{Binary Convolution}
After converting the input to binary image, XNOR convolution is applied on the binary image. A simple iterative XNOR operation is enough to obtain XNOR convolution outputs. The pseudo-code is given in Algorithm \ref{alg:xnor}.
\begin{algorithm}
  \begin{algorithmic}[1]
  \STATE $Mask = weight\_matrix\_of\_ones$
  \STATE $iteration\_row = register\_size_x - kernel\_size_x + 1$
  \STATE $iteration\_col = register\_size_y - kernel\_size_y + 1$
    \FOR{$j < iteration\_row$}
      \FOR{$i < iteration\_col$}
        \STATE $register\_image = (input\_image[j][i] >> Shift) \oplus Weight\_Kernel $
        \STATE $register\_image = register\_image \land Mask $
        \STATE $Shift += 1$\\
      \ENDFOR
    \ENDFOR
  \end{algorithmic}
   \caption{XNOR Convolution}
   \label{alg:xnor}
\end{algorithm}
The theoretical speed-up that can be achieved for this part is 58x for $1\times1$ kernel size \cite{xnor}. However, networks used in computer vision use larger kernel sizes to have a receptive field, hence convolutions with larger kernels are needed in practice. In this work, we use a kernel size of $3\times3$ which results in a more modest speed-up as it necessitates an iterative approach.  In \cite{xnor}, convolution kernel weights fill every bit inside a register. For $1\times1 $ kernel size, this involves copying the same sign value for each bit in register.  When $3\times3$ convolution kernels are used, XNOR convolution can not be applied to each of the bit-pixel value since bits at the edge of the registers will require padding. Hence in our implementation, the weight register only contains one meaningful weight value and the other $Register\_image\_size - kernel\_size$ bits are masked by bitwise AND operation.
\subsubsection{Binary Image to Integer Image}
Output of the XNOR convolution is still in binary image format and each convolution result is stored inside a single register. To convert the convolution result into integer, the total number of 1-bits in the register needs to be counted. 

In our implementation, in order to count the 1-bits inside the registers, we use the relevant x86\_64 instruction and the special function provided by CUDA (\verb#__popc#) in CPU and GPU implementations respectively. 
\subsubsection{Multiplication by Scaling Factor}
\label{subsec:ScalingFactor}

By averaging $\mathbf{X} \in \mathbb{R}^{c \times w \times h }$ across channels, $\mathbf{A} \in \mathbb{R}^{w \times h}$ is obtained. $\mathbf{A}$ is convolved with a matrix $\mathbf{k}$ to get scaling factor matrix $\mathbf{K} \in \mathbb{R}^{w \times h}$, which is multiplied with output.

We implemented multi-threaded versions of both vanilla convolution and XNOR convolution on CPU as baseline methods to compare against the parallelized versions on GPU. In this section, we first describe the CPU implementation. This is followed by the description of GPU implementation. 


\subsection{Binary Convolution on CPU}

The CPU implementation has the following steps. 
\begin{enumerate}[wide, labelwidth=!, labelindent=0pt]
    \item Apply zero padding to the Tensor (3D).
    \item Convert the tensor and weights to binary type.
    \item Apply bit-wise XNOR operation on binary Tensor.
    \item Convert binary Tensor to integer Tensor.
    \item Repeat steps 2, 3, 4 for all input channels and sum the results across input channels.
    \item Repeat steps 2, 3, 4, 5 for output channels (filters).
    \item Summation on input Tensor across channels to find scaling factor.
    \item Scalar Multiplication of output result from (5) with $\alpha$ and scaling factor.
\end{enumerate}

\subsubsection{Converting Integer to Binary}
The input and image tensors are stored inside registers with \verb#unsigned long# data type to fully utilize 64-bit CPU registers and benefit from 64-bit operations. Each 64-bit register can hold 64 data elements of a $8\times8$ matrix. Hence, the input image is divided into $8\times8$ tiles, each of which is then stored in a single register in binary form. The pseudo-code is provided in Algorithm \ref{alg:inputtobinary}. 
\subsubsection{Binary Convolution}
In this part, CPU registers are used since there is up to $36\times$ iterative access to the same register. The pseudo-code is given in Algorithm \ref{alg:xnor}.
\subsubsection{Binary Image to Integer Image}
Each convolution result is stored inside a single 64-bit register. To convert the convolution result into integer value, the total number of 1s in the register need to be counted. This can be done by using the special built-in function \verb#__builtin_popcount# of the GCC compiler, which performs this operation more efficiently than hash mapping.

\subsubsection{Multiplication by Scaling Factor}
This part is done as described in \ref{subsec:ScalingFactor}.
\subsection{Binary Convolution on GPU}
In  convolution, XNOR operation and scaling factor calculation are independent, hence they can be run asynchronously in two different CUDA streams. XNOR convolution result is then obtained by multiplication of the outputs of these two streams.
The GPU implementation has the following steps running in two different streams:\\
Stream 1:
    \begin{enumerate}[wide, labelwidth=!, labelindent=10pt]
    \item Apply zero padding to the Tensor (3D).
    \item Convert the tensor and weights to binary type.
    \item Apply bit-wise XNOR operation on binary Tensor.
    \item Convert binary Tensor to integer Tensor.
    \end{enumerate}
Stream 2:
    \begin{enumerate}[wide, labelwidth=!, labelindent=10pt]
        \item Summation on input Tensor across channels to find scaling factor.
        \item Scalar Multiplication of  $\alpha$ and scaling factor.
    \end{enumerate}


\subsubsection{Input Scaling Factor}

For calculating the scaling factor of input, the average of channels is taken and convolved with 2D filter $\mathbf{k} \in \mathbb{R}^{w \times h}$ as in Eq. (\ref{compK}). It includes 2 steps: averaging across channels and convolution. For summation, each thread calculates the sum of a pixel across channels. 

Computing input scaling factor matrix includes following steps after copying from host to device:
\begin{enumerate}
    \item Set grid and block sizes.
    \item Average pixels across input channels.
    \item Execute memory specified CUDA function to compute kernel convolution.
    \item Deallocate GPU memories.
\end{enumerate}
\subsubsection{Binary Convolution Operation} 
Main idea of algorithm is similar to the CPU version. However, while CPU registers are 64-bit, GPU registers are 32-bit and each register now holds a $(8\times4)$ image tile rather than $(8\times8)$ tiles. Each CUDA thread converts a $(8\times4)$ image tile (stored in a single register) to binary in parallel. The total number of threads to  launch can be calculated as in Eq. \ref{total_t}, where $I_{W}$, $I_{H}$, $R_{W}$, $R_{H}$, $K_{W}$, $K_{H}$, represents the input image width, image height, register width, register height, kernel width and kernel height respectively. 


\begin{equation}
    \label{total_t}
    Total_{t} = \frac{I_{W} - R_{W}}{R_{W} + 1 - K_{W}} \times \frac{I_{H} - R_{H}}{R_{H} + 1 - K_{H}}
\end{equation}

After converting the image batches to binary, XNOR convolution is applied on all input channels. Denoting the number of input channels and number of output channels by $\#in\_ch$ and $\#out\_ch$ respectively, a total of $\#out\_ch$ different convolutions (having different weights) are calculated for each input channel. For this purpose, two options were explored: (\textit{i}) using a single kernel to calculate all $\#out\_ch$ binary convolutions on all input, (\textit{ii}) using $\#out\_ch$ number of kernels calculating binary convolution for each output channel separately. The first approach results in better utilization as it uses the register space for the whole process without any need for copying the result to global memory. However, this approach prohibits parallelization on output channels. For each input channel, $\#out\_ch$ convolution operations needs to be calculated and these operations are executed by the same kernel thread iterating a loop for the $\#out\_ch$ times. Therefore, the second approach preferred. In that case, since conversion of integer image to binary image can be stored in global memory, multiple streams can access these data. As a result $\#out\_ch$ streams can be run asynchronously, resulting in better paralellization.

\subsubsection{Multiplication by Scaling Factor}
A straight forward multiplication of $convolution\_result \times K \times \alpha $ for each CUDA thread. 
\section{Experimental Evaluation}
\begin{table}[t]
\centering
\caption{Comparison of vanilla convolution with XNOR convolution on GPU (ms).}
\label{tab:comp_gpu}
\begin{tabular}{|c|c|c|c|}
\hline
\textbf{Input Size}  & \textbf{Vanilla Conv.} & \textbf{XNOR Conv.} & \textbf{Speed-up} \\ \hline
\textbf{$256\times256$}     & 0.062                            & 0.024             & $2.57\times$ \\ \hline
\textbf{$512\times512$}     & 0.186                            & 0.069             & $2.69\times$ \\ \hline
\textbf{$1024\times1024$}   & 0.671                            & 0.252             & $2.66\times$ \\ \hline
\textbf{$2048\times2048$}   & 2.641                            & 0.986             & $2.68\times$ \\ \hline
\end{tabular}
\end{table}
\begin{table}[t]
\centering
\caption{Comparison of CPU and GPU performance for vanilla convolution (ms).}
\label{tab:conv_gpu_cpu}
\begin{tabular}{|c|c|c|c|}
\hline
\textbf{Input Size} & \textbf{CPU} & \textbf{GPU} & \textbf{Speed-up} \\ \hline
\textbf{$256\times256$}        & 3.437           & 0.061           & $56.34\times$ \\ \hline
\textbf{$512\times512$}        & 10.623             & 0.186           & $57.11\times$ \\ \hline
\textbf{$1024\times1024$}       & 35.811             & 0.671          & $53.37\times$ \\ \hline
\textbf{$2048\times2048$}       & 132.714              & 2.641           & $50.25\times$ \\ \hline
\end{tabular}
\end{table}
\begin{table}[t]
\centering
\caption{Comparison of CPU and GPU performance for XNOR convolution (ms).}
\label{tab:xnor_gpu_cpu}
\begin{tabular}{|c|c|c|c|}
\hline
\textbf{Input Size} & \textbf{CPU} & \textbf{GPU} & \textbf{Speed-up} \\ \hline
\textbf{$256\times256$}        & 0.743           & 0.0237           & $31.35\times$ \\ \hline
\textbf{$512\times512$}        & 2.531            & 0.0692           & $36.57\times$ \\ \hline
\textbf{$1024\times1024$}       & 10.088           & 0.2519           & $40.04\times$ \\ \hline
\textbf{$2048\times2048$}       & 42.011          & 0.9859            & $42.61\times$ \\ \hline
\end{tabular}
\end{table}

We have run the experiments on a system having Intel i7700 CPU with 4-cores and Nvidia GTX1080TI GPU. The time measurements take only the computation into account so memory operations like allocation, copying memory and deallocation are excluded. For multi-core CPU implementation, OpenMP has been used. The sub parts that are explained above are made for single channel input matrices. 
For the GPU implementation, CUDA has been used. We observed that the performance was insensitive to block size, so the experiments have been conducted with a constant block size of 256.  We have used a constant $3\times3$ kernel size throughout the experiments for both CPU and GPU versions. All the experiments have been repeated 100 times and average run-times have been calculated. 

As shown in Table \ref{tab:comp_gpu}, GPU XNOR convolution provides a speed-up of $2.57\times$ to $2.69\times$ against GPU vanilla convolution.  the speed-up remains fairly constant with different image sizes.

When CPU and GPU implementations are compared, it is observed that vanilla convolution has a speed-up of $50.25\times$ to $57.11\times$ on GPU (Table \ref{tab:conv_gpu_cpu}). XNOR convolution has a speed-up of $31.35\times$ to $42.61\times$ (Table \ref{tab:xnor_gpu_cpu}) and speed-up increases with increasing input size due to better utilization of the GPU. 

\section{Discussion}

 While the GPU XNOR convolution implementation has better performance than the CPU XNOR and GPU vanilla counterparts, the speed-ups we observed were lower than those reported in \cite{xnor}. It has to be noted that our design uses a single kernel for each logical operation and as such, lacks the ability to achieve $32\times$ (assuming 32-bit registers) binary logical operation speed. So, further optimizations could leverage bit-wise parallelism. On the other hand, use of separate registers allows easier conversion from binary outputs to integers.  
 XNOR convolution needs binarization of input and multiplication with scaling factor at the end. Converting integer input image values to binary values and restoring integer values from output of the XNOR convolution are costly operations as they require sequential write operation to modify each bit inside a register and read them after convolution. For a deeper network, this process may optimized by passing the binary outputs to the next kernel without integer conversion.

XNOR convolution involve two processes that can run concurrently, which are computing scaling matrix $\mathbf{K}$ and binary convolution operation. As a future work, multiple streams can be used to overlap these operations.





\section{Conclusions}

We have implemented and optimized the XNOR convolution operation \cite{xnor} used in binary convolutional networks on CPU and GPU and comparatively evaluated their performance. The experimental results show that up to $42.61\times$ speed-up can be achieved on GPU compared to the multi-threaded CPU implementation. 

We implemented the operations required for the whole inference path of the binary network (i.e. scaling factor calculation and multiplication, binary to integer and integer to binary conversion, XNOR convolution) and made the code publicly available at \url{https://github.com/metcan/Binary-Convolutional-Neural-Network-Inference-on-GPU}. However it has to be noted that the operations other than XNOR convolution part are not optimized and developed for testing only. Hence, for a real-life deployment requiring high levels of performance, these parts also need to be optimized.

\addtolength{\textheight}{-12cm}   




\bibliography{references}

\end{document}